\documentclass[conference]{IEEEtran}
\IEEEoverridecommandlockouts
\usepackage{cite}
\usepackage{amsmath,amssymb,amsfonts}
\usepackage{algorithmic}
\usepackage{graphicx}
\usepackage{textcomp}
\usepackage{url}
\usepackage{xcolor}
\def\BibTeX{{\rm B\kern-.05em{\sc i\kern-.025em b}\kern-.08em
    T\kern-.1667em\lower.7ex\hbox{E}\kern-.125emX}}
\begin{document}

\title{FloodLense: A Framework for ChatGPT-based Real-time Flood Detection
}

\author{\IEEEauthorblockN{Pranath Reddy Kumbam and Kshitij Maruti Vejre}
\IEEEauthorblockA{\textit{University of Florida, Gainesville, USA}\\
}}

\maketitle

\begin{abstract}
This study addresses the vital issue of real-time flood detection and management. It innovatively combines advanced deep learning models with Large language models (LLM), enhancing flood monitoring and response capabilities. This approach addresses the limitations of current methods by offering a more accurate, versatile, user-friendly and accessible solution. The integration of UNet, RDN, and ViT models with natural language processing significantly improves flood area detection in diverse environments, including using aerial and satellite imagery. The experimental evaluation demonstrates the models' efficacy in accurately identifying and mapping flood zones, showcasing the project's potential in transforming environmental monitoring and disaster management fields.
\end{abstract}


\section{Introduction}

\subsection{Societal Applications}

The increasing prevalence and severity of flood events, exacerbated by climate change and rapid urbanization, underscore the critical need for effective flood detection and management strategies. Climate change contributes to more frequent and intense weather phenomena, leading to unpredictable and devastating floods. Urbanization further aggravates this issue by creating impermeable surfaces that hinder natural water absorption and drainage, elevating flood risks in densely populated areas. These challenges are compounded by the substantial impact of floods on societies, economies, and the environment. Flooding disasters result in significant loss of life, displacement of communities, extensive damage to infrastructure, and long-lasting economic disruptions. Additionally, they pose severe environmental threats, including ecosystem destruction and water pollution. Current flood management practices often struggle to keep pace with these challenges. Traditional methods face limitations in real-time data acquisition, coverage, and predictive accuracy, highlighting the urgent need for more advanced, integrated, and scalable solutions. This research aims to address these gaps by introducing a novel flood detection system, leveraging the latest advancements in deep learning and natural language processing to offer a more effective approach to flood monitoring and response. For example, during sudden flash floods, users in a vulnerable area could query their locality, receive a real-time visual update, and make informed decisions regarding evacuation or safety precautions. 

\subsection{Existing Methods}

Image processing techniques have played a significant role in flood detection. Techniques such as edge detection, pixel analysis, and the utilization of remote sensing technologies have been successful \cite{8955273} \cite{10.1007/978-981-13-3708-6_32} \cite{Bijeesh2020}. These approaches require manual feature engineering, and their inherent lack of scalability, adaptability, and robustness have motivated a shift towards data-driven machine learning models.

Some of the earlier data-driven approaches relied heavily on using thresholding-based approaches on hand-crafted features to discriminate between floodwater and confusers like vegetation and shadows \cite{nhess-11-529-2011}. These traditional methodologies typically required domain experts to hand-craft features, which could be a time-consuming and expertise-driven process. The authors in \cite{w14071140} employed traditional machine learning algorithms like Random Forest (RF), Support Vector Machine (SVM), and Maximum Likelihood Classifier (MLC) to process satellite imagery for detecting flooded areas. Regression analysis and tree-based approaches have also been employed for predicting heavy rain damage \cite{SNEHIL202078}. 

The authors in \cite{w14071140} have also implemented a novel unsupervised machine learning framework called the change detection approach based on a combination of the Otsu algorithm, fuzzy rules, and iso-clustering methods for urban flood detection and reported a superior performance compared to traditional supervised methods. The use of change detection has been extensively explored in some of the other works \cite{6297453}. While an unsupervised machine learning algorithm such as change detection is more robust due to its higher speed, lesser requirements for training data, and computation runtime, some of the more traditional change detection implementations such as the one used by \cite{6297453} are highly prone to excess false positives.    

Traditional supervised methods can achieve high accuracy given a sufficiently large and well-labeled training dataset. However, They require a significant amount of labeled data, which can be time-consuming and expensive to obtain. The automatic identification of features through deep learning algorithms can reduce the reliance on manual feature engineering and domain expertise, thus potentially improving the accuracy, efficiency, and timeliness of flood detection from satellite imagery.

In recent years, deep learning models have significantly impacted the realm of flood segmentation and detection \cite{2018arXiv180509757X}, often outperforming the traditional methods outlined in section 1.1. Models such as U-Net and Convolutional Neural Networks (CNNs) \cite{10.1007/978-981-19-7524-0_23} \cite{rs15082046} \cite{10.1007/978-3-030-92600-7_16} \cite{LI2023161757} have risen to prominence due to their ability to learn complex patterns from large datasets without the necessity for manual feature engineering. Furthermore, the application of modified U-Net architectures and other tailored deep learning models like FAPNET have also been explored for enhancing flood-water detection in satellite remote sensing applications \cite{s22218245}. 

LSTM, a type of recurrent neural network which has shown promise in time-series prediction, has been utilized for time-series prediction which is relevant for flood prediction \cite{Zhong2023} \cite{FANG2021125734}. The intersection of spatial and temporal data dynamics is crucial for a comprehensive understanding and prediction of flood events. To that extent, Spatial-Temporal Graph Deep Learning Models \cite{Farahmand2023} have also been explored for flood nowcasting, focusing on spatial and temporal data dynamics. Furthermore, unsupervised and semi supervised approaches \cite{2022arXiv221203675Y} \cite{2021arXiv210708369P} have been harnessed to overcome the challenges associated with data labeling, showcasing potential for automating flood detection and segmentation from satellite imagery without the necessity for extensive labeled data.

The increased application of deep learning models in flood segmentation and detection underscores the transition towards more data-driven, automated, and precise approaches in flood management, leveraging the availability of large datasets and advanced computational techniques to augment the identification and analysis of flood events from satellite imagery. Deep learning methods can handle complex data structures and are capable of learning features automatically from the data, potentially achieving higher accuracy compared to traditional machine learning methods. However, they require a large amount of labeled data and computational resources. The availability of high-quality flood datasets is rather limited, which can hinder the training of deep learning models, limiting their ability to generalize to new terrains, geographical, and climatic and other real-world scenarios. While the computational complexity of deep learning algorithms is an inherent problem, we aim to solve the later shortcoming with our approach that leverages the generalization ability \cite{2023arXiv230514712Y} of a generative conditional diffusion model.

Diffusion models are an emerging class of deep generative models that reconstruct high-fidelity samples through iterative denoising \cite{2020arXiv200611239H}. Recent works have applied conditional diffusion models to diverse image-to-image translation tasks like segmentation and inpainting \cite{2021arXiv211200390A} \cite{2022arXiv220109865L}, demonstrating photorealistic output. Diffusion models have also been used for medical image segmentation \cite{2022arXiv221100611W}. Such conditional generative modeling is promising for flood detection and visualization but remains unexplored.

This section focuses on the review of some of the existing interfaces for remote sensing and flood detection. While flood detection has benefited enormously from remote sensing and computer vision advances, advances in flood detection algorithms must be accompanied by progress in public accessibility. Below are some of the organizations that offer static interfaces for remote sensing and flood detection.

The USGS employs WaterWatch \cite{usgs_waterwatch_2023}, collects flood data through the USGS's extensive network of streamgages. It doesn't harness satellite imagery hence it can’t be used for post-flood analysis. Limitation is that it can’t be used to find the regions actually affected by the flood.

The Global Flood Detection System from JRC \cite{kugler2007global} uses the capabilities of AMSR-E's 36 GHz H-polarization data to detect daily riverine inundation events. The methodology, however, struggles in areas with intricate land use patterns or topography, occasionally making it difficult to differentiate between wet soils, water bodies, and genuine floodwaters. Another limitation is its reliance on historical baselines that, if not updated, can sometimes lead to inaccuracies in flood predictions.

MODIS NRT Global Flood Product \cite{nasa_modis_nrt_flood_2023} from NASA LANCE represents another significant stride in flood detection. This technology capitalizes on the Modified Normalized Difference Water Index (MNDWI) to ascertain flood-affected regions. While effective, it faces challenges comparable to JRC's system, especially in regions with complex topographies and diverse land use patterns \cite{Li2022}.

On the algorithmic front, GEE has championed a unique approach to identify flood-affected terrains. By using the threshold detection technique grounded in Otsu's Algorithm, it seeks to provide reliable flood data. However, its results sometimes grapple with the issue of misclassification, thereby limiting its accuracy in certain scenarios \cite{google_earth_engine_project_2023}.

Lastly, MAXAR's Open Data Program \cite{maxar_open_data_2023} endeavors to provide a visual representation of flood-affected zones. By furnishing high-resolution before-and-after images, it presents a stark visual comparison. However, it restricts itself to imagery without delving into deeper analytical aspects, thus serving more as a visualization tool than a comprehensive flood analysis interface.

Static interfaces including the traditional graphical user interface (GUI) based systems for flood detection and monitoring pose significant challenges around limited accessibility and usability for non-experts including the general public. Traditional interfaces can be difficult to use since they require the input query to be formulated precisely through parameters, coordinates, etc. The lack of interactivity and contextual feedback in static interfaces makes it difficult for users to explore data interactively or get clarifications on model outputs. Moreover, non-conversational systems cannot handle follow-up questions. Integrating a flood detection backend with conversational interfaces such as GPT \cite{openai_research_2023} could help overcome these limitations by enabling natural language conversations that make flood detection/monitoring more contextual and accessible.

\subsection{Technical Contributions and Project Novelty}

This project introduces significant technical advancements in flood detection by integrating state-of-the-art deep learning models with the ChatGPT platform. Key contributions include:

\begin{itemize}
  \item \textbf{Integration:} The combination of UNet, Residual Dense Network (RDN), and Vision Transformer (ViT) models with ChatGPT represents a novel approach, enhancing the system's capability to process and interpret complex environmental data effectively.
  \item \textbf{Advanced Image Processing:} The application of these deep learning models for processing aerial and satellite imagery significantly improves flood area detection and mapping accuracy, outperforming traditional methods.
  \item \textbf{User Interaction:} By leveraging ChatGPT's natural language processing capabilities, the system offers an intuitive and user-friendly interface, allowing users to interact and receive real-time flood data in a conversational manner.
  \item \textbf{Contribution to Disaster Management:} This project represents a significant step forward in environmental monitoring and disaster response, providing a scalable and efficient tool for timely flood detection and management.
\end{itemize}

These contributions not only demonstrate the project's technical prowess but also highlight its potential to revolutionize flood management practices, offering a more responsive, accurate, and user-centric solution.

\subsection{Limitations of Current Approaches
}

Current state-of-the-art flood detection approaches \cite{2018arXiv180509757X} \cite{10.1007/978-981-19-7524-0_23} \cite{rs15082046} \cite{10.1007/978-3-030-92600-7_16} \cite{LI2023161757} are hindered by several machine learning challenges that limit their practical application. Traditional models are heavily data-dependent, necessitating extensive, well-labeled datasets that are both costly and time-consuming to assemble, which impedes their ability to generalize across diverse flooding events. These models often rely on manual feature engineering, requiring significant domain expertise, and may still fail to capture the multifaceted nature of floods, leading to inaccuracies. Furthermore, the models' applicability is constrained by computational demands and a lack of interpretability, which is crucial in emergency situations where understanding the rationale behind predictions is vital. Compounding these issues is the absence of practical, user-friendly applications that can translate complex model outputs into actionable insights for disaster response. Most existing solutions do not cater to non-expert users and lack real-time processing capabilities, which are essential during time-critical events. Additionally, there is a notable gap in integrating these models with existing emergency management systems to provide a seamless end-to-end response mechanism. Our project seeks to bridge these gaps by creating an accessible, real-time flood detection tool that non-technical users can easily interact with, thus enhancing the practical utility of flood detection in urgent scenarios.

The introduction of a conditional diffusion model represents a significant innovation in the field. This model type utilizes an iterative generative process to refine predictions, providing a powerful mechanism for capturing the complex patterns associated with flood delineation. Unlike traditional deep learning models, the diffusion model has the potential to produce high-fidelity segmentation maps with fewer artifacts and more precise boundaries. Additionally, the project explores the potential of Vision Transformers (ViT), which use self-attention mechanisms to process satellite images. This allows the model to focus on the most relevant parts of an image, improving the accuracy of flood detection, particularly in complex environments where traditional CNNs may falter. By integrating these models with ChatGPT, FloodLense provides an intuitive, natural language-based interface that allows users to request flood information simply by asking questions. This ease of use represents a significant step forward in making advanced flood detection technology accessible to a broader audience, including those without technical expertise.

\section{Problem definition}

The central problem addressed in this project is the real-time detection and mapping of flood areas using advanced deep learning techniques integrated with a conversational AI platform. The primary components of this problem include:

\begin{enumerate}
    \item \textbf{Inputs:} The system inputs are high-resolution aerial and satellite images, which capture diverse landscapes under various weather conditions and urban developments. These images are critical in identifying potential flood-affected areas.
    \item \textbf{Deep Learning Models:} The system utilizes a combination of UNet, Residual Dense Network (RDN), and Vision Transformer (ViT) for image analysis. UNet is pivotal for precise localization in the images, essential for delineating flood boundaries accurately. RDN and ViT are integrated to enhance image details and pattern recognition. RDN focuses on improving image resolution and clarity, crucial in differentiating flooded areas from other water bodies or wet surfaces. ViT, on the other hand, processes images as sequences of patches, providing a comprehensive view of large-scale environmental patterns, enhancing the detection of widespread flooding.
    \item \textbf{Output:} Accurate identification and mapping of flooded regions, presented in a format that is both informative and accessible to a range of users, from disaster management professionals to the general public.
    \item \textbf{Objective:} The main goal is to develop a robust, scalable, and efficient flood detection system capable of rapidly analyzing complex environmental imagery and providing actionable, real-time insights.
\end{enumerate}

Consider a scenario where a Coastal city is facing imminent flooding due to prolonged heavy rain or rise in the sea levels. For example we want to check the food situation in Chennai, We can ask the GPT to finding the current flood situation in Chennai and The system requests real-time satellite imagery from SentinalHub and passes it as input. Utilizing the combined strengths of UNet, RDN, and ViT, the system accurately processes these images to identify and segment flood-affected areas. The processed data is then relayed through the ChatGPT interface, offering users detailed information about the flood's impact. This facilitates prompt and efficient response actions by disaster management teams, potentially saving lives and reducing property damage.

\begin{figure}
    \centering
    \includegraphics[width=1\linewidth]{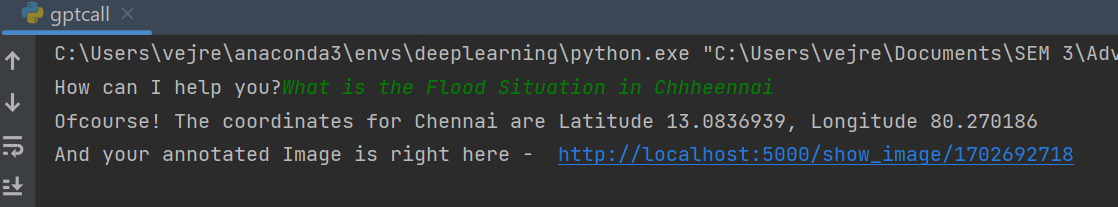}
    \caption{ChatGPT plugin Simulation}
    \label{fig:fig1}
\end{figure}

\begin{figure}
    \centering
    \includegraphics[width=1\linewidth]{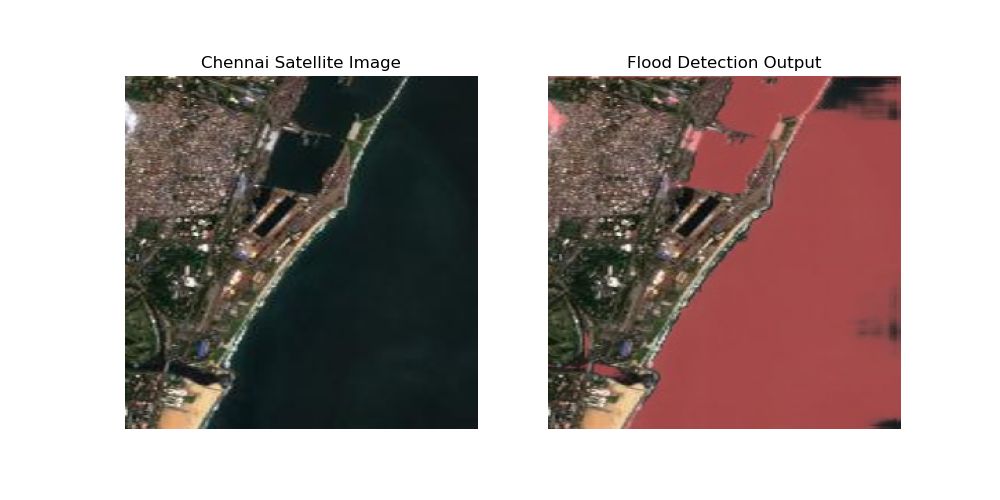}
    \caption{Original vs Highlighted Flood Detected Image}
    \label{fig:fig2}
\end{figure}

In Figure \ref{fig:fig1}. We can see that even though the prompt is not currently providing the city name "What is the Flood Situation in Chhheennai", yet it provides the correct image url to view the flood situation in Chennai, India. Figure \ref{fig:fig2}. shows the original image and the flood affected highlighted areas. 

\section{Proposed solution}

The FloodLense project introduces a comprehensive solution to flood detection that harnesses the power of advanced deep learning techniques and user-friendly interfaces. The central proposition is to utilize a blend of neural network architectures, including U-Net \cite{2015arXiv150504597R}, Residual Dense Network (RDN) \cite{2018arXiv181210477Z}, Vision Transformer (ViT) \cite{2020arXiv201011929D}, and conditional diffusion models \cite{2020arXiv200611239H}, to analyze and segment satellite imagery for real-time flood monitoring.

The U-Net architecture, specifically designed for biomedical image segmentation, features a symmetric encoder-decoder structure. The encoder path follows the typical convolutional neural network (CNN) architecture, systematically reducing the spatial dimensions while increasing the depth to capture contextual information. Conversely, the decoder path up-samples the feature maps and combines them with high-resolution features from the encoder via skip connections, essential for precise localization and segmentation of spatial structures in satellite imagery.

The Residual Dense Network (RDN) enhances this by employing densely connected layers. Each layer in RDN receives inputs from all preceding layers, fostering feature reuse and propagation, and improving the learning of detailed representations. This is particularly effective in differentiating subtle features in complex images, such as distinguishing water from land in flooded areas.

Vision Transformers (ViTs) diverge from traditional CNNs by applying self-attention mechanisms, allowing the model to weigh different parts of the input image differently. This approach enables ViTs to capture long-range dependencies and global context in the image, which is instrumental in understanding extensive and diverse flood patterns. Unlike CNNs, which process local receptive fields, ViTs can analyze the image in a more holistic manner, making them well-suited for large-scale environmental pattern recognition.

\begin{figure}[h]
    \centering
    \includegraphics[width=0.9\linewidth]{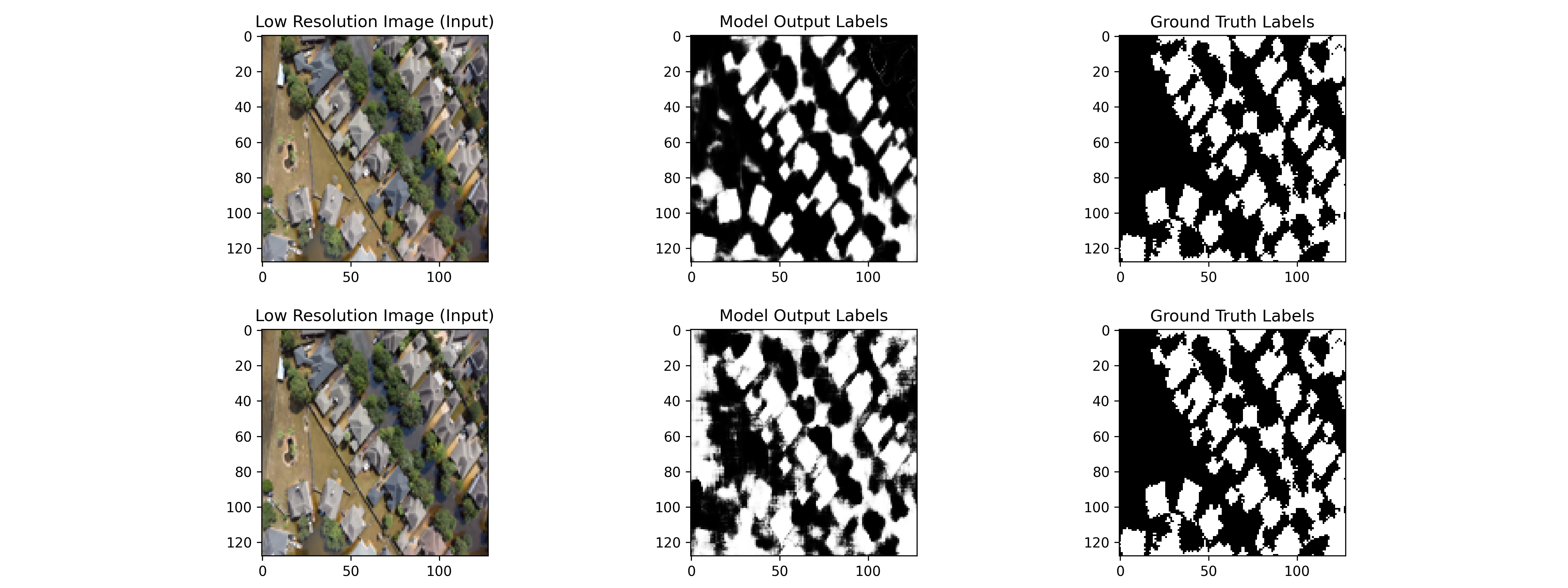}
    \caption{Sample Output Showing Flooded Buildings - UNet (Top), RDN (Bottom)}
    \label{fig:fig3}
\end{figure}

Together, these models provide a comprehensive approach to flood detection, combining local precision with global contextual understanding, which is critical for accurate and efficient flood mapping. Moreover, we plan to explore the capabilities of conditional diffusion models, a class of generative models that have demonstrated success in creating high-fidelity images through an iterative process of denoising. These models hold the promise of generating highly detailed and accurate segmentation maps, essential for identifying flood-affected areas. In the preliminary analysis, we have explored a U-Net and Residual Dense Network models using the FloodNet dataset \cite{9460988} \cite{rahnemoonfar2020floodnet}. The results produced are visually promising as the segmentation map presented in Figure \ref{fig:fig3} shows, indicative of the models' potential in flood detection.

\begin{figure}[h]
    \centering
    \includegraphics[width=1.0\linewidth]{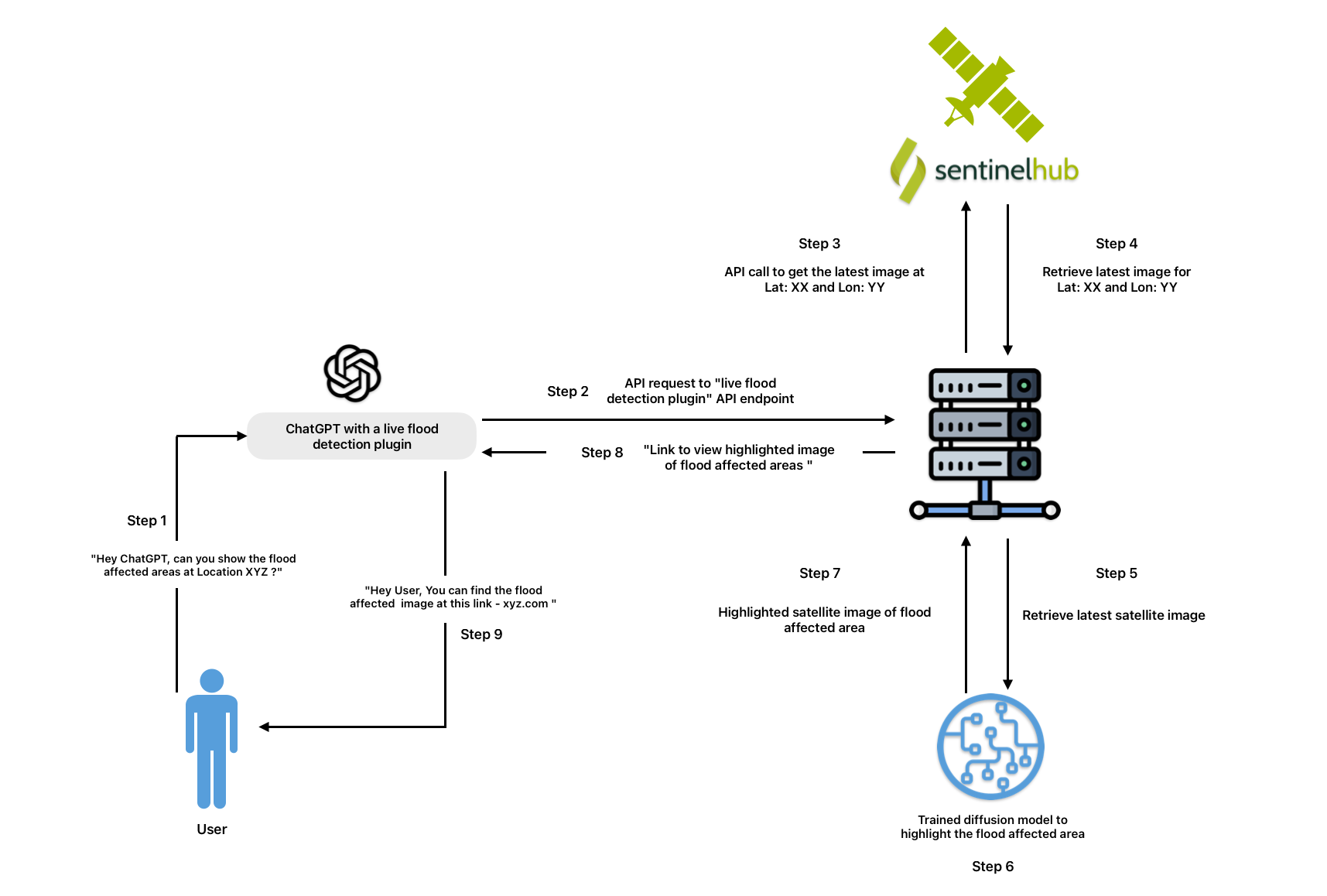}
    \caption{System Architecture}
    \label{fig:fig4}
\end{figure}

To make the system as shown in Figure \ref{fig:fig4} accessible to users without technical expertise, we plan to integrate ChatGPT \cite{chatgpt2023}, which will act as an interface between the user and the deep learning models. Users will be able to input queries in natural language, which ChatGPT will interpret to extract relevant geographical and temporal data. This interaction simplifies the user experience, making the process of accessing flood information simple.

SentinelHub \cite{sentinelhub_engine} will be used as the primary source of satellite data, providing up-to-date and high-resolution imagery necessary for the models to perform segmentation. We will develop an API that will fetch the required images from SentinelHub in response to user queries processed by ChatGPT. The API will be designed for efficiency.

We have developed a Flask-based web application designed to provide access to satellite imagery via a simple user interface. The application, hosted on Replit, allows users to retrieve and display Sentinel-2 satellite images based on specified latitude and longitude coordinates. A key feature of this application is the /download\_image endpoint, which dynamically generates a bounding box around the provided coordinates and retrieves the latest true-color image from the Sentinel-2 data collection. This image is then saved in a server directory, with its URL returned to the user for access.

\subsection{Psuedo-Code for Satellite Image Processing and Retrieval}

\begin{enumerate}
    \item \textbf{Initialize Configuration}
    \begin{itemize}
        \item Set instance ID for Sentinel Hub Configuration.
    \end{itemize}
    
    \item \textbf{Define Function to Retrieve Coordinates}
    \begin{itemize}
        \item \textit{Input:} Request parameters.
        \item \textit{Process:} Retrieve latitude and longitude from request, with default values.
        \item \textit{Output:} Bounding box coordinates based on input latitude and longitude.
    \end{itemize}
    
    \item \textbf{Define Function to Retrieve and Process Satellite Data}
    \begin{itemize}
        \item \textit{Input:} Bounding box coordinates.
        \item \textit{Process:}
        \begin{enumerate}
            \item Request Sentinel-2 data for the bounding box.
            \item Select the latest image from the data.
            \item Process the image:
            \begin{itemize}
                \item Crop, resize, and convert color format.
            \end{itemize}
            \item Save the image with a timestamp in a designated directory.
            \item Log image information.
            \item Generate a URL for accessing the image.
        \end{enumerate}
        \item \textit{Output:} JSON response with the image URL.
    \end{itemize}
\end{enumerate}

\subsection{Psuedo-Code to Simulate ChatGPT Plugin}

\begin{enumerate}
    \item \textbf{Initialize OpenAI Client}
    \begin{itemize}
        \item Set API key.
    \end{itemize}
    
    \item \textbf{Define Custom Tool for Location Retrieval}
    \begin{itemize}
        \item Tool to extract location from user input.
    \end{itemize}
    
    \item \textbf{Define Function to Get Location Details}
    \begin{itemize}
        \item \textit{Input:} Location query in natural language.
        \item \textit{Process:} Use OpenAI API to interpret the location query.
        \item \textit{Output:} Extracted location details.
    \end{itemize}
    
    \item \textbf{Define Function to Geocode Location}
    \begin{itemize}
        \item \textit{Input:} Location name.
        \item \textit{Process:} Use Nominatim service to convert location name to coordinates.
        \item \textit{Output:} Latitude and longitude of the location.
    \end{itemize}
    
    \item \textbf{Define Function to Get Satellite Data}
    \begin{itemize}
        \item \textit{Input:} Latitude and longitude.
        \item \textit{Process:}
        \begin{enumerate}
            \item Send a request to a local server with coordinates.
            \item Retrieve the URL of the processed image.
        \end{enumerate}
        \item \textit{Output:} Display the URL or error message.
    \end{itemize}
    
    \item \textbf{Main Function}
    \begin{itemize}
        \item \textit{Process:}
        \begin{enumerate}
            \item Prompt user for location query.
            \item Extract location details using OpenAI's chat completion.
            \item Geocode the location to get coordinates.
            \item If coordinates are found, display them and retrieve the satellite image.
            \item If not, display an error message.
        \end{enumerate}
    \end{itemize}
\end{enumerate}

\section{Evaluation}

This study embarks on a critical evaluation of various models for flood and water detection. The study's cornerstone lies in its exploration of the capability of deep learning models such as UNet  \cite{2015arXiv150504597R}, Residual Dense Net (RDN) \cite{2018arXiv181210477Z}, and Vision Transformer (ViT) \cite{2020arXiv201011929D} in accurately segmenting and identifying water bodies in diverse imaging scenarios. The investigation navigates through the intricacies of model performance, computational demands, and the pragmatic viability of these models in real-world applications. Additionally, we also assess the effectiveness of the interface application designed to extract location details from user-input sentences related to flood and water detection.

Specifically, we will be investigating the following questions:
\begin{itemize}
  \item How do different models like UNet, Residual Dense Net (RDN), and Vision Transformer (ViT) compare in terms of segmentation accuracy?
  \item What are the implications of model complexity on computational efficiency and the interpretability of results?
  \item How does sensitivity to threshold variations impact model performance, particularly with respect to false positives?
  \item How accurate is the extraction of locations from natural language inputs?
  \item How reliable is the process of converting these extracted locations into geographic coordinates?
  \item What is the model's capability in handling a variety of location-related queries, including those that are ambiguous or refer to non-existent locations?
\end{itemize}

\subsection{Evaluation Data and Experimental Configuration}

In this study, two carefully chosen datasets form the backbone of the experimental evaluation: the FloodNet aerial drone dataset \cite{9460988} \cite{rahnemoonfar2020floodnet} for preliminary model assessment, and the Sentinel satellite dataset \cite{sentinelhub_engine} for training the final model selected for deployment. This strategic delineation ensures a comprehensive analysis of model capabilities in diverse imaging scenarios, crucial for flood detection applications.

The FloodNet dataset is used for initial comparative analysis of the deep learning models. It consists of aerial drone images that capture a multitude of flood situations, providing a rich tapestry of data for understanding the challenges in aerial-based flood detection. The dataset, tailored for supervised flood detection tasks, encompasses 2,108 training samples, ensuring a varied learning environment for the models. For the purpose of evaluating the models' accuracy in segmentation, 235 testing samples are included. The images have been downscaled to a resolution of 128x128 pixels using nearest-neighbor interpolation. This method is specifically chosen for its capability to resize images with minimal interpolation errors, thus preserving the integrity of the original data. Each image is a 3-channel RGB image and has undergone normalization to standardize the input for the models, eliminating variations due to disparate lighting conditions or scales.

The Sentinel satellite dataset, with its focus on satellite imagery of water bodies, serves as the real-world application context for the final model selected post-preliminary analysis. This dataset is composed of 2,820 training samples, offering a different perspective and scale for water body detection from space. The dataset is rounded off with 41 testing samples, essential for assessing the model's practical deployment capabilities. The images are reshaped and downscaled to a resolution of 256x256 pixels, again using nearest-neighbor interpolation to maintain the fidelity of the satellite imagery while making it amenable to deep learning processing. Like the FloodNet dataset, these images are also 3-channel and normalized to maintain preprocessing consistency.

The selection of UNet, RDN, and ViT models for this study is influenced by their varied architectural designs, which facilitates a broad spectrum analysis across different deep learning methodologies. An additional model, a diffusion model, was initially considered. However, due to its substantial computational complexity, the study was constrained to working with images of merely 64x64 dimensions when using this model, a scale that is suboptimal for effective flood detection analysis. Consequently, the focus shifted to the other three models, which offered a more feasible balance between computational demand and performance efficacy. The preprocessing steps, including normalization and resizing using nearest-neighbor interpolation, ensure that the data across both datasets is uniformly processed, setting a standard baseline for model evaluation.

The test data for our study consisted of sentences designed to simulate real-world queries concerning floods and water bodies, incorporating a variety of geographical references such as specific cities, regions, rivers, and landmarks. This dataset featured a range of location types, including names of cities, rivers, and notable landmarks, to ensure a comprehensive evaluation. Our model employed OpenAI's GPT-3.5-turbo \cite{chatgpt2023} for the natural language processing aspect, effectively interpreting and extracting location information from the queries. For the conversion of these textual location descriptors into geographical coordinates, we integrated Nominatim's geocoding service, enhancing the model's capability to accurately geocode the extracted data.

\subsection{Evaluation Metric}

In assessing the performance of deep learning models for flood/water detection, a comprehensive set of evaluation metrics is employed to quantify their capabilities and practical applicability. Key metrics include the Intersection over Union (IoU) and the Dice Coefficient, both of which are essential in segmentation tasks. IoU, or the Jaccard Index, measures the overlap between the predicted segmentation and the ground truth, indicating the model's accuracy in identifying flood-affected areas. The Dice Coefficient, related to IoU, assesses segmentation efficiency, especially useful in imbalanced classes typical in flood detection. These metrics provide a nuanced view of the models' precision in segmenting areas accurately, a critical aspect in flood detection applications.

Complementing these are metrics like Precision, Recall, and the F1 Score. Precision calculates the proportion of correctly predicted positive observations, while Recall measures the proportion of actual positives correctly identified, both crucial in evaluating the model's effectiveness in identifying water regions and flood-affected areas. The F1 Score, the harmonic mean of Precision and Recall, offers a balanced measure, particularly vital in scenarios where a trade-off between these two metrics might occur. Additionally, Accuracy gives an overview of the model’s overall performance, and Inference Time indicates the model's computational efficiency, essential in time-sensitive flood detection scenarios. Lastly, the Sensitivity to Threshold Variations metric is significant for understanding the models’ robustness and their ability to maintain accuracy under different operational conditions, making these metrics collectively pivotal in offering a comprehensive evaluation of the models' segmentation accuracy, overall performance, and practical applicability in diverse real-world scenarios.

In evaluating the ChatGPT integration's performance for location extraction and geocoding, we are utilizing three key metrics. First, 'Accuracy' gauges the precision in identifying and extracting location names from textual input. Secondly, the 'Geocoding Success Rate' measures the proportion of accurately geocoded locations, reflecting the system's ability to accurately map extracted names to geographical coordinates. Lastly, the 'Error Rate' is assessed, which captures the frequency of erroneous geocoding outcomes or instances where the system fails to correctly identify a location name. These metrics collectively provide a comprehensive overview of the system's efficacy in handling geocoding tasks within natural language processing contexts.

\subsection{Performance Analysis and Results}

The study's exploration into the capabilities of UNet, Residual Dense Net (RDN), and Vision Transformer (ViT) models across FloodNet and Sentinel datasets provides an insightful dissection of their performance in flood detection tasks. This section delves deeper into the individual and comparative analyses of these models, focusing on their performance across various metrics.

In the preliminary analysis using the FloodNet dataset, the UNet model's performance emerged as a key focal point. As seen in Figure \ref{fig:fig3}, UNet's segmentation outputs were observed to be markedly superior, suggesting a nuanced understanding and handling of the complex spatial features characteristic of flood-affected areas. Despite this visual proficiency, there was a noticeable disparity when compared to its quantitative metrics as seen in \ref{table:metrics}. This variance underscores a crucial aspect of model evaluation in practical applications: the potential limitations of standard quantitative metrics in capturing the true effectiveness of a model in complex real-world scenarios. UNet's ability to outperform in visual segmentation accuracy, despite not leading in quantitative metrics, hints at its architectural strengths in effectively processing and delineating complex aerial images, a vital attribute for accurate flood detection.

\begin{figure}[h]
    \centering
    \includegraphics[width=1.0\linewidth]{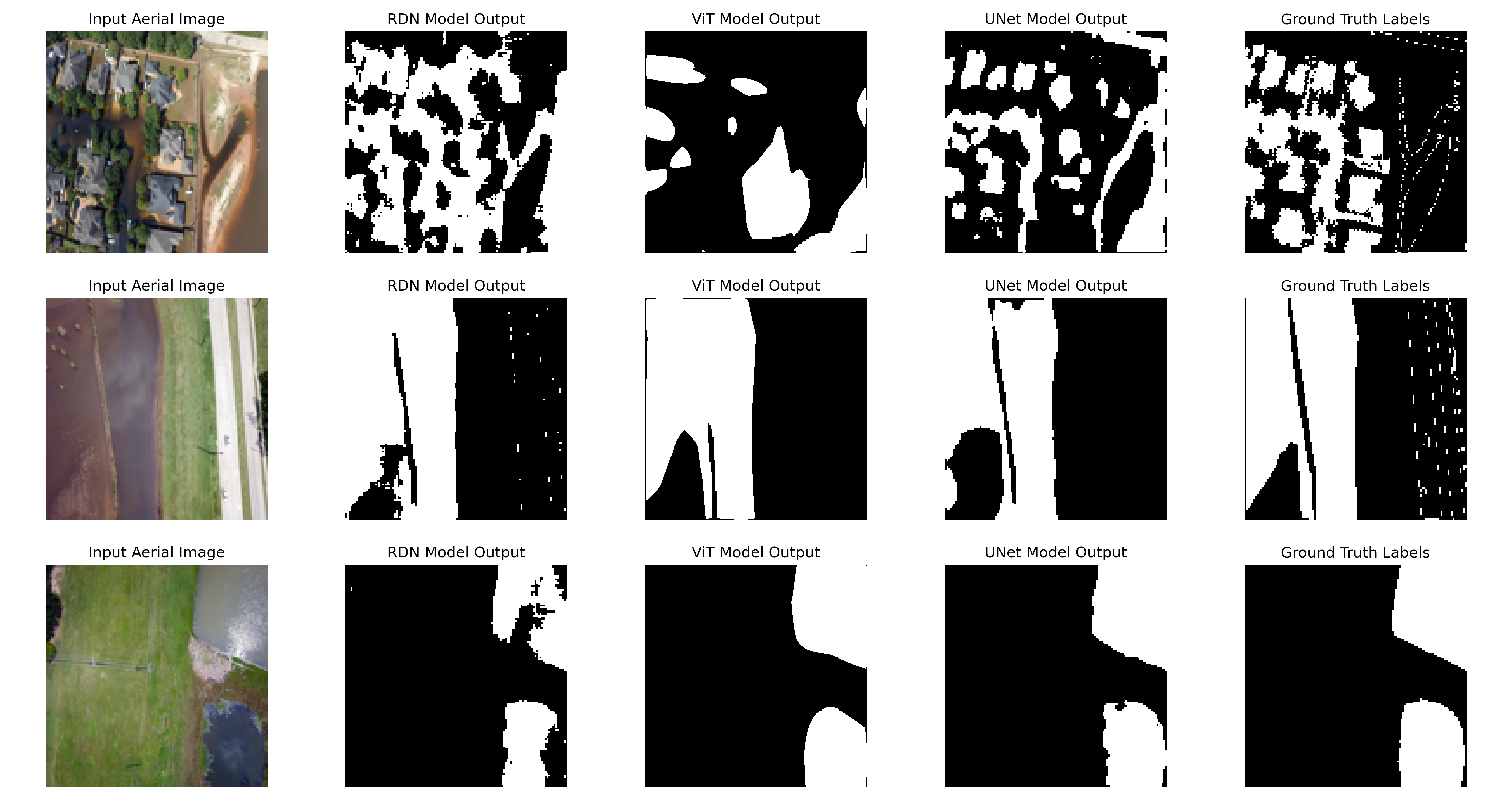}
    \caption{Sample Outputs from FloodNet Data. Positive values indicate water and flooded regions.}
    \label{fig:fig5}
\end{figure}

\begin{table}[h!]
\centering
\caption{Performance Metrics}
\label{table:metrics}
\begin{tabular}{|c|c|c|c|}
\hline
Metric & RDN & ViT & UNet \\
\hline
IoU & 0.49412 & 0.45401 & 0.42273 \\
Dice & 0.56414 & 0.52523 & 0.47872 \\
Precision & 0.57131 & 0.68377 & 0.57269 \\
Recall & 0.59470 & 0.51801 & 0.49207 \\
F1 Score & 0.56414 & 0.52523 & 0.47872 \\
Accuracy & 0.90969 & 0.89435 & 0.90674 \\
\hline
\end{tabular}

\end{table}

\begin{table}[h!]
\centering
\caption{Inference Time on FloodNet Data}
\label{table:inference_time}
\begin{tabular}{|c|c|c|c|}
\hline
Model & RDN & ViT & UNet \\
\hline
Inference Time (ms) & 10.41157 & 66.04217 & 10.70199 \\
\hline
\end{tabular}

\end{table}

The evaluation of the RDN model in the FloodNet dataset context brought to light its capabilities and limitations in flood scenario segmentation. While the RDN model exhibited high levels of segmentation accuracy, it slightly trailed behind the UNet model. This finding is significant as it indicates that although RDN is proficient in handling the diverse range of features presented in flood scenarios, it may not be as adept as UNet in capturing certain finer details. This slight difference in performance could be attributed to the inherent architectural differences between the two models, with RDN possibly requiring further tuning to achieve the same level of detail recognition as UNet in complex flood imagery.

The performance of the ViT model on the FloodNet dataset revealed a notable gap between its theoretical capabilities and practical output. Despite being designed as an advanced architecture, ViT's segmentation outputs were less accurate when compared visually. This gap between its quantitative metrics and visual performance raises questions about the model's suitability for detailed segmentation tasks like those required in flood detection. It highlights the potential misalignment between ViT’s architecture and the specific requirements of fine-grained, localized segmentation tasks, a critical consideration for its application in flood scenario analysis.

In the ablation study conducted on the FloodNet dataset, significant insights were gleaned into the internal structure and critical components of the Residual Dense Net (RDN), Vision Transformer (ViT), and UNet models. For the RDN model, the ablation of different blocks and layers led to varying degrees of decline in performance metrics such as precision, recall, and F1 score. This variability in the impact of ablation across different layers underscored the nuanced role each component plays in the model's overall performance, highlighting the importance of certain layers more than others in accurate flood detection.

The study revealed a stark dependency of the ViT model on its decoder layers, with metrics dropping to zero upon their ablation, indicating these layers' crucial role in segmentation tasks. In contrast, the UNet model exhibited significant performance drops, even resulting in undefined values in some cases when specific encoding layers were removed. This suggests a critical reliance on these layers for the model’s functionality in flood detection, pointing to their role in effective feature processing. These findings from the ablation study are instrumental in understanding the architectural strengths and vulnerabilities of these models, providing a foundation for future enhancements in their design for improved flood detection capabilities. The complete ablation study results have been presented in the Appendix.

\begin{figure}[h]
    \centering
    \includegraphics[width=1.0\linewidth]{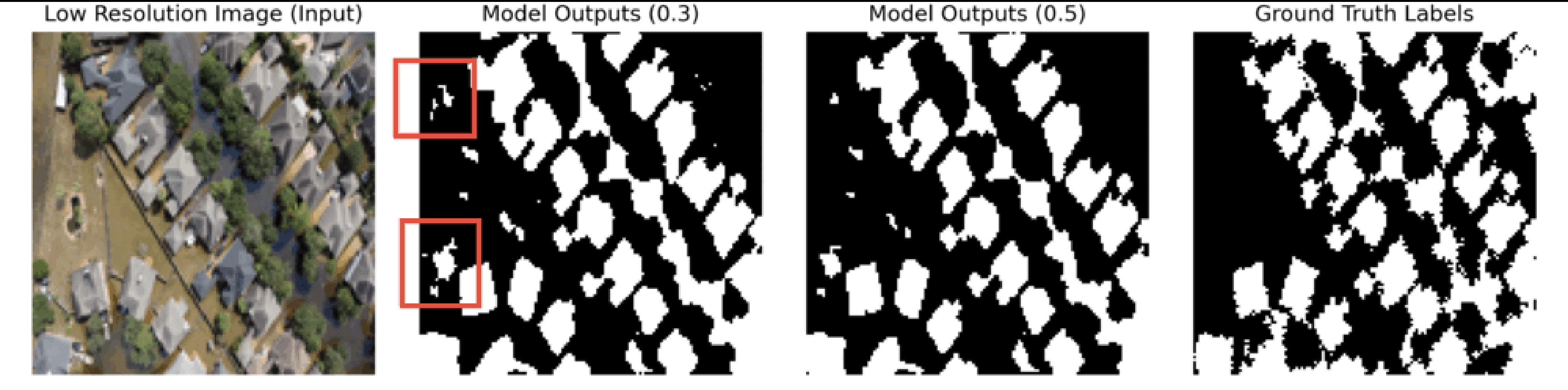}
    \caption{Samples for Multiple Output Thresholds. The false positives identified in the lower threshold have been highlighted by red squares.}
    \label{fig:fig6}
\end{figure}

The sensitivity of these models to threshold variations was another critical aspect of this study. Specifically, the increase in false positives at lower threshold levels, such as 0.3, pointed to a trade-off between sensitivity and precision. This phenomenon was particularly observable in the UNet model, where a lower threshold led to the model being overly sensitive, erroneously identifying certain non-flooded regions as flooded. This sensitivity is a crucial operational factor, as it necessitates the careful calibration of threshold settings to achieve an optimal balance for accurate flood detection. The complete matrics across various thresholds have been presented in the Appendix.

The decision to focus on the UNet model for the Sentinel dataset analysis was influenced by its superior performance in the preliminary FloodNet dataset evaluation. UNet's ability to accurately segment flood-affected areas, as evidenced by its visual outputs, along with its computational efficiency as seen in \ref{table:inference_time}, positioned it as a promising model for further testing and application in the different context of satellite imagery provided by the Sentinel dataset.

\begin{figure}[h]
    \centering
    \includegraphics[width=1.0\linewidth]{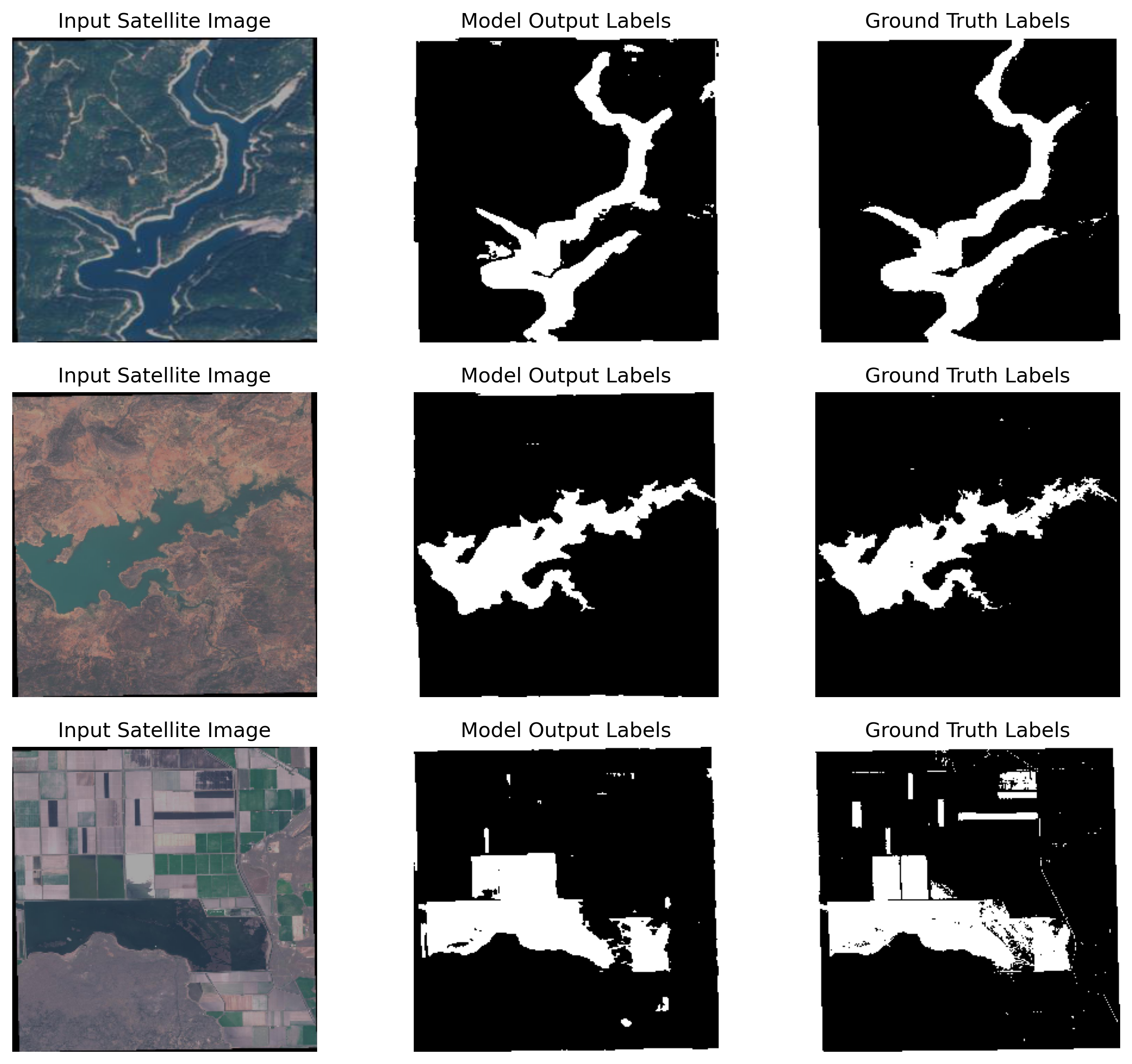}
    \caption{Sample Outputs from Sentinel Data. Positive values indicate water and flooded regions.}
    \label{fig:fig7}
\end{figure}

\begin{table}[h!]
\caption{Performance Metrics on Sentinel Data (UNet)}
\label{table:performance_metrics_sentinel}
\centering
\begin{tabular}{|c|c|}
\hline
Metric & Value \\
\hline
IoU & 0.65117 \\
Dice & 0.76080 \\
Precision & 0.77475 \\
Recall & 0.78513 \\
F1 Score & 0.76080 \\
Accuracy & 0.87711 \\
\hline
\end{tabular}
\end{table}

Upon analyzing the UNet model with the Sentinel dataset, a clear improvement was noted across all quantitative metrics as seen in \ref{table:performance_metrics_sentinel}. This enhancement indicated not only the model's adaptability but also its effectiveness in a new imaging context. However, the slight increase in inference time underscored the necessity for models to be optimized for varying data resolutions and complexities, as encountered in satellite versus aerial imagery. Furthermore, the ablation studies on the Sentinel dataset revealed potential vulnerabilities in the UNet model's architecture, specifically in instances where undefined values were observed. This aspect of the analysis highlighted areas within the model's architecture that could be optimized for enhanced performance and reliability in real-world applications. Lastly, the interpretability analysis through activation maps provided deeper insights into the model's processing behavior and response to satellite imagery, indicating potential areas for performance enhancement and refinement. 

A notable observation was the sparsity of activations in the encoder layers when processing the Sentinel satellite imagery. This sparsity indicates that the model, in its current architecture, selectively focuses on specific features or regions within the satellite images. While this focused approach suggests efficiency in identifying key features relevant to flood detection, it also raises the possibility that the model might be overlooking other potentially significant information. 

\begin{enumerate}
    \item \textbf{Visualization:} No visualization provided, as the output is textual.
    \item \textbf{Quantitative Metrics:} The model showed high accuracy in standard scenarios but struggled with ambiguous and non-existent locations. The geocoding service successfully converted most locations to coordinates, except for fictional or ambiguous ones.
    \item \textbf{Case Study:} For the sentence "Tsunami alerts for the coast of Japan", the model returned "Japan" instead of "Coast of Japan", indicating a limitation in handling descriptive locations. In the case of "Weather forecast for Mount Everest", the model incorrectly identified "Mount Everest" as a valid location for flood detection, showing a need for better filtering of irrelevant queries.
    \item \textbf{Analysis:} The model performs well with clear, specific locations but has limitations in understanding more complex or descriptive queries. The inability to handle non-existent locations like "Atlantis" indicates a need for a validation layer before geocoding. The high success rate in standard scenarios shows promise for practical application, with scope for refinement in handling edge cases.
\end{enumerate}

In an extensive evaluation of deep learning models including UNet, Residual Dense Net (RDN), and Vision Transformer (ViT) across FloodNet and Sentinel datasets, key insights have been gleaned regarding their efficacy in flood detection. UNet's performance, in particular, stood out with its superior visual segmentation accuracy across both datasets, affirming its effectiveness in delineating flood-affected areas. This was notable despite some discrepancies between visual and quantitative data, underscoring UNet's potential as a reliable tool for real-world flood detection. Conversely, ViT, while advanced in design, did not meet expectations in segmentation accuracy, indicating a need for alignment between model architecture and the specific requirements of flood detection tasks.

The study also highlighted the sensitivity of these models to threshold settings, where a higher rate of false positives at lower thresholds was particularly evident in the UNet model. This finding emphasizes the importance of fine-tuning threshold settings to balance sensitivity and precision, which is crucial for practical applications. Recommendations from the study include optimizing UNet for broader deployment, carefully calibrating thresholds to suit environmental conditions, and improving real-time processing for time-sensitive scenarios. Additionally, exploring hybrid models that combine strengths of different architectures, diversifying datasets, and collaborating with interdisciplinary experts are suggested to enhance the models' applicability and relevance in disaster management.

Finally, the interface model demonstrated proficiency in extracting and geocoding locations from straightforward natural language inputs, but faced few challenges with descriptive or ambiguous queries and in filtering out irrelevant or fictional locations. This underscores the need for future development to focus on enhancing the models' understanding of complex queries and incorporating validation mechanisms. Such advancements are critical for improving the models' functionality in diverse real-world flood and water detection scenarios, meeting the dynamic challenges in environmental monitoring and disaster management. This comprehensive study thus provides a roadmap for refining these deep learning models, ensuring they are more attuned to the nuanced requirements of flood detection and response.

\section{Conclusion}

In conclusion, this research represents a significant stride in flood management technology. By leveraging advanced deep learning models in tandem with ChatGPT's conversational AI, this project has successfully addressed some of the most pressing challenges in real-time flood detection and mapping. The system's ability to analyze complex environmental data through aerial and satellite imagery and provide accurate, timely flood information is a noteworthy accomplishment.

This research addresses a critical need in the context of increasing climate change impacts and urbanization. The integration of deep learning models such as UNet with ChatGPT enhances the accuracy and efficiency of flood detection over traditional methods, which often struggle with real-time data processing and broad coverage. Furthermore, the user-friendly interface of ChatGPT opens up this technology to a wider range of users, from disaster management professionals to the general public.

The project's experimental results demonstrate not only the feasibility of such an integrated approach but also its potential to revolutionize the field of environmental monitoring and disaster response. The system's scalability and adaptability to different environments and conditions underscore its practicality and applicability in various geographical settings.

Looking ahead, future research can explore further enhancements to the system, such as integrating additional environmental factors, improving prediction accuracy, and expanding geographical coverage. The potential to adapt this technology for other environmental monitoring applications, such as wildfire detection or air quality monitoring, also presents exciting avenues for further development.

This research underscores the vital role of technological innovation in addressing environmental challenges. By advancing flood detection capabilities, this project contributes significantly to improving disaster preparedness and response, ultimately aiding in safeguarding communities and ecosystems against the growing threat of floods.

\section*{Acknowledgment}

We would like to thank Dr. Zhe Jiang for his guidance and the University of
Florida for providing access to educational resources.

\newpage
\bibliographystyle{unsrt}
\bibliography{bibo}

\newpage
\section{Appendix}

\subsection{Preliminary Results :On Aerial Drone Flood Images}

\begin{table}[h!]
\centering
\caption{UNet Performance Metrics}
\label{table:unet}
\begin{tabular}{|c|c|c|c|c|c|}
\hline
Metric & Threshold 0.3 & 0.4 & 0.5 & 0.6 & 0.7 \\
\hline
IoU & 0.43301 & 0.42650 & 0.42273 & 0.41835 & 0.41302 \\
Dice & 0.49098 & 0.48274 & 0.47872 & 0.47429 & 0.46890 \\
Precision & 0.57501 & 0.56790 & 0.57269 & 0.57329 & 0.57311 \\
Recall & 0.51038 & 0.49959 & 0.49207 & 0.48446 & 0.47623 \\
F1 Score & 0.49098 & 0.48274 & 0.47872 & 0.47429 & 0.46890 \\
Accuracy & 0.90683 & 0.90672 & 0.90674 & 0.90644 & 0.90575 \\
\hline
\end{tabular}

\end{table}

\begin{table}[h!]
\centering
\caption{Residual Dense Net Performance Metrics}
\label{table:residual_dense_net}
\begin{tabular}{|c|c|c|c|c|c|}
\hline
Metric & Threshold 0.3 & 0.4 & 0.5 & 0.6 & 0.7 \\
\hline
IoU & 0.49718 & 0.49627 & 0.49412 & 0.49215 & 0.48899 \\
Dice & 0.56819 & 0.56676 & 0.56414 & 0.56174 & 0.55899 \\
Precision & 0.56014 & 0.56697 & 0.57131 & 0.57582 & 0.58092 \\
Recall & 0.61566 & 0.60475 & 0.59470 & 0.58495 & 0.57423 \\
F1 Score & 0.56819 & 0.56676 & 0.56414 & 0.56174 & 0.55899 \\
Accuracy & 0.90709 & 0.90868 & 0.90969 & 0.91063 & 0.91081 \\
\hline
\end{tabular}

\end{table}

\begin{table}[h!]
\centering
\caption{Vision Transformer Performance Metrics}
\label{table:vision_transformer}
\begin{tabular}{|c|c|c|c|c|c|}
\hline
Metric & Threshold 0.3 & 0.4 & 0.5 & 0.6 & 0.7 \\
\hline
IoU & 0.46295 & 0.45869 & 0.45401 & 0.44926 & 0.44506 \\
Dice & 0.53512 & 0.53039 & 0.52523 & 0.51969 & 0.51530 \\
Precision & 0.67567 & 0.67880 & 0.68377 & 0.68568 & 0.63724 \\
Recall & 0.53205 & 0.52533 & 0.51801 & 0.51019 & 0.50293 \\
F1 Score & 0.53512 & 0.53039 & 0.52523 & 0.51969 & 0.51530 \\
Accuracy & 0.89621 & 0.89532 & 0.89435 & 0.89332 & 0.89238 \\
\hline
\end{tabular}

\end{table}

\begin{table}[h!]
\centering
\caption{Ablation Study Results for RDN, ViT, and UNet Models on FloodNet Dataset}
\label{tab:ablation_study}
\begin{tabular}{|c|c|c|c|c|}
\hline
\textbf{Model} & \textbf{Ablated Part} & \textbf{Precision} & \textbf{Recall} & \textbf{F1-Score} \\
\hline
RDN & Block 0 Layer 0 & 0.5113 & 0.5812 & 0.5239 \\
RDN & Block 0 Layer 1 & 0.5090 & 0.5842 & 0.5276 \\
RDN & Block 0 Layer 2 & 0.5136 & 0.5541 & 0.5120 \\
RDN & Block 1 Layer 0 & 0.4476 & 0.5686 & 0.4639 \\
RDN & Block 1 Layer 1 & 0.4438 & 0.5040 & 0.4382 \\
RDN & Block 1 Layer 2 & 0.4220 & 0.5580 & 0.4425 \\
\hline
ViT & Decoder Layers 0,1,2,3 & 0.0 & 0.0 & 0.0 \\
\hline
UNet & enc1 & 0.1842 & 0.8 & 0.2719 \\
\hline
\end{tabular}
\end{table}

\end{document}